\documentclass{article}

\PassOptionsToPackage{numbers, compress}{natbib}
\usepackage[preprint]{neurips_2026}

\usepackage[utf8]{inputenc} 
\usepackage[T1]{fontenc}    
\usepackage{hyperref}       
\usepackage{url}            
\usepackage{booktabs}       
\usepackage{amsfonts}       
\usepackage{nicefrac}       
\usepackage{microtype}      
\usepackage{xcolor}         
\definecolor{darkgreen}{RGB}{0,100,0}

\usepackage{graphicx}
\usepackage{subcaption}

\usepackage{amsmath} 
\usepackage{amssymb}
\usepackage{multirow} 
\usepackage{wrapfig}

\usepackage{makecell}
\usepackage{pifont}  

\newcommand{\cmark}{\ding{51}}
\newcommand{\xmark}{\ding{55}}

\title{World-Coherent Decoding: Self-Verifying Test-Time Planning for World Action Models}

\author{%
  Chuhan Zhang\\
  Department of Computer Science\\
  Institute of Science Tokyo\\
  \texttt{chuhan\_mio@d-itlab.comp.isct.ac.jp}
  \And
  Seiji Ito\\
  Department of Computer Science\\
  Institute of Science Tokyo\\
  \texttt{ito.s.0948@m.isct.ac.jp}
  \And
  Kenta Hoshino\\
  Department of Computer Science\\
  Institute of Science Tokyo\\
  DENSO IT Lab, Japan\\
  \texttt{hoshino@comp.isct.ac.jp}
  \And
  Satoshi Ikehata\\
  National Institute of Informatics, Japan\\
  DENSO IT Lab, Japan\\
  \texttt{ikehata.satoshi@d-itlab.comp.isct.ac.jp}
  \And
  Ikuro Sato\\
  Department of Computer Science\\
  Institute of Science Tokyo\\
  DENSO IT Lab, Japan\\
  \texttt{isato@comp.isct.ac.jp}
}

\begin{document}

\maketitle

\providecommand{\methodname}{WCD}

\begin{abstract}
World Action Models (WAMs) aim to control robots by stochastically generating visual futures and then decoding actions, but empirical observations indicate that the results can strongly depend on which future is selected. 
We propose World-Coherent-Decoding (WCD), a self-verifying test-time planning framework that treats WAM rollouts as falsifiable future--action hypotheses. 
At each decision step, WCD samples multiple candidates from a frozen WAM and ranks them using internal generative signals: flow-based video surprisal for visual plausibility and action path effort for action-generation stability. 
After execution, the realized observation audits the selected imagination, yielding an imagination--reality mismatch that trains a lightweight online predictor for future candidate selection. 
Thus, WCD converts delayed self-verification into pre-execution reliability estimation without updating the backbone model. 
On RoboTwin 2.0, WCD improves Hard success under limited randomized-scene supervision from $55.80\%$ to $60.90\%$, with a $+16.43$ gains on Horizon-3 tasks, and shows qualitative robustness on real Franka visual-shift tests. 
These results highlight a simple principle: test-time scaling for WAMs depends less on sampling more futures than on selecting reliable ones.
\end{abstract}

\section{Introduction}
World Action Models (WAMs)~\cite{li2026causal, ye2026worldactionmodelszeroshot, kim2026cosmos} have recently emerged as a promising alternative to purely reactive vision-language-action (VLAs) policies~\cite{kim2024openvla,black2024pi0,black2025pi05}. 
Unlike VLAs that directly map observations and language instructions to actions, WAMs incorporate video-based world modeling to predict how the visual world may evolve under interaction, internalizing physical priors from large-scale video data: how objects move, how scenes deform and how manipulation sequences unfold accordingly.
Among them, \emph{causal}~\cite{li2026causal,ye2026gigaworld,hu2025video} video-action WAMs such as LingBot-VA~\cite{li2026causal} jointly model visual futures and action chunks in an autoregressive structure, explicitly tying imagined future dynamics to the actions that produce them. 

This future-conditioned action decoding can be viewed through an inverse-dynamics
lens~\cite{tian2024predictive,du2023learning,hu2025video}: the predicted future visual state
acts as an implicit goal, and the action branch decodes the motion required to reach it.
This structure connects classical goal-conditioned control with modern pixel- and latent-space visual dynamics~\cite{fragkiadaki2015learning,srinivas2018universal}. 
More importantly for test-time control, it exposes candidate visual consequences before physical commitment, making causal WAMs especially appealing under distribution shift or temporally extended manipulation~\cite{ebert2018visual}.

\begin{figure}[t]
    \centering
    \includegraphics[width=\linewidth]{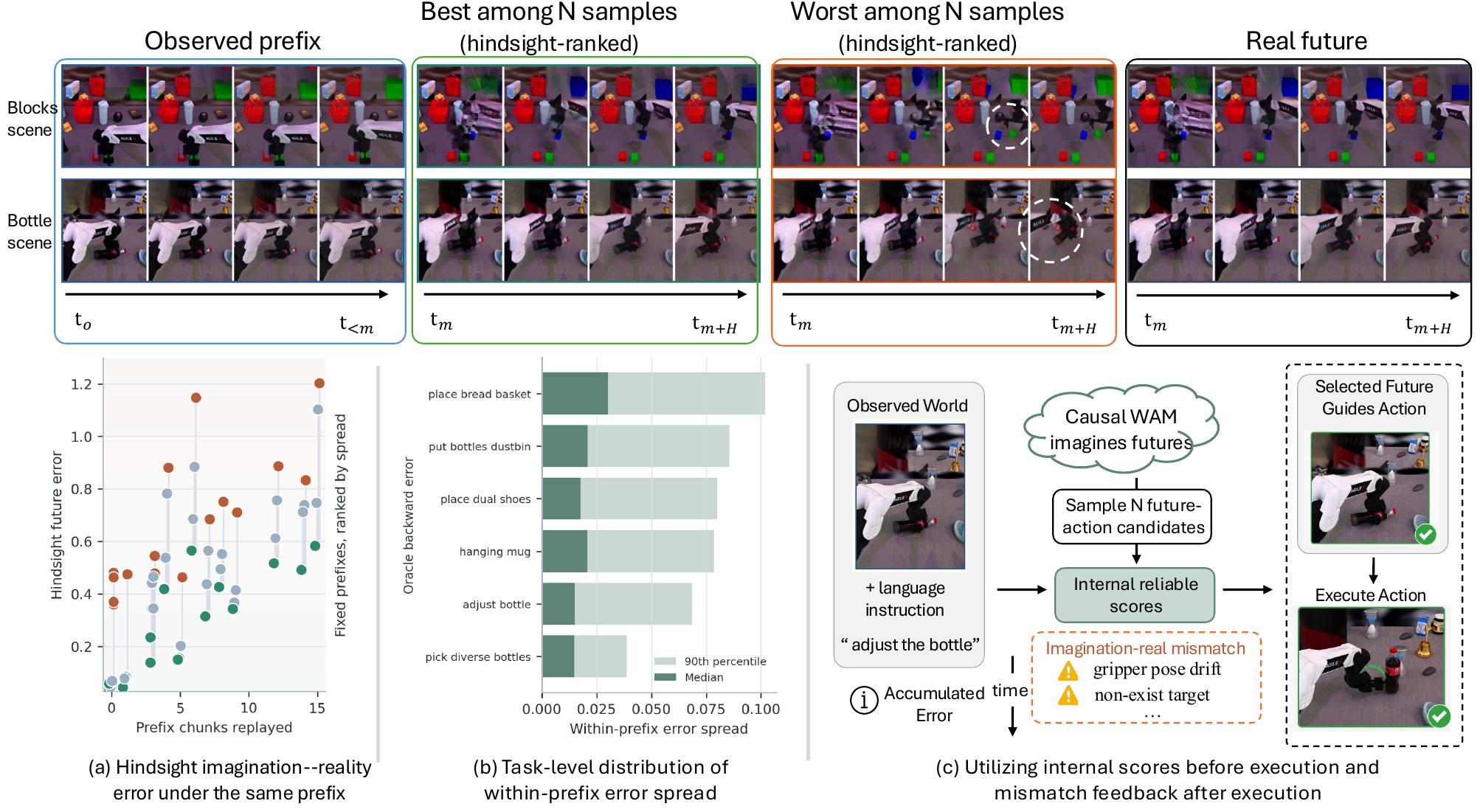}
    \caption{
    \textbf{Pilot motivation: WAM imagination contains selectable reliability variation.}
    A frozen WAM samples multiple future--action candidates from the same causal history, and each
    imagined visual future is scored in hindsight by its latent-space imagination--reality error after the
    real future becomes available.
    (a) Under the same observation prefix, candidates can differ sharply in hindsight error; green marks
    the lowest-error candidate, brown the highest, and gray the remainder.
    (b) The within-prefix spread appears across tasks, measured by the median and 90th-percentile
    oracle error spread over prefixes.
    (c) Reliability-aware candidate selection is applied before execution, and self-supervised mismatch feedback is given after execution.
    Implementation details are provided in Appendix~\ref{appendix:pilot}.
    }
    \vspace{-16pt}
    \label{fig:hook_pilot}
\end{figure}

However, realizing this potential is non-trivial, because video generation is inherently stochastic.
Under the same observation prefix and task instruction, a WAM can sample futures of widely different quality: some remain physically consistent and task-relevant, while others introduce nonexistent objects of implausible object poses (see the top row of Figure~\ref{fig:hook_pilot}) that make consistent action decoding difficult.
These observations motivate us to investigate a selection rule for test-time imaginations so that the WAM can more successfully perform a desired control.

The similar issues are broadly found in stochastic generation, where random sampling alone does not guarantee downstream utility and candidate ranking or refinement is often needed for reliable deployment~\cite{kirstain2023pick,xu2023imagereward,li2025reflect,eyring2025noise}.
For WAM-based control, this reliability problem is more consequential: an imagined future is not merely a generated artifact, but an action-conditioned hypothesis whose reliability directly affects physical execution.

\vspace{-16pt}
\paragraph{Pilot: WAM imagination contains selectable reliability variation.}
We first ask whether stochastic WAM futures contain enough within-prefix variation so that candidate selection is reasonable. We replay fixed observation prefixes into the frozen WAM cache, sample multiple future--action candidates from the same causal history, and evaluate each imagined visual future in hindsight after the corresponding real future becomes available.

Figure~\ref{fig:hook_pilot} (a) shows a clear reliability spread within the sampled pool.
Even with the same observation history and instruction, different stochastic futures can yield substantially different imagination--reality errors, and this effect appears across multiple tasks (Figure~\ref{fig:hook_pilot} (b)). Thus, test-time imagination is not merely noisy. Qualitatively, high-error futures can correspond to recognizable failure patterns, such as nonexistent
target objects, implausible object motion, or gripper-pose drift, rather than only small latent-space deviations; it contains a selectable gap between a one-shot sample and better candidates in the same pool. 
Complementarily, online rollouts show that larger early imagination--reality mismatch is associated with higher failure rate (Figure~\ref{fig:error_success_failure_connection}), suggesting that this mismatch is behaviorally meaningful rather than only a latent-space diagnostic.

This leads to the core idea behind our proposed method: WAM imagination is \emph{falsifiable}.
A WAM predicts an explicit future before execution, and the realized observation can later audit that prediction. The resulting mismatch provides a self-supervised reliability signal without task rewards, success detectors, or external trained verifiers (Figure~\ref{fig:hook_pilot} (c)). The difficulty is temporal: this signal is revealed only after execution, while the action must be selected beforehand.

In this work, we propose World-Coherent Decoding (\methodname{}): Self-Verifying Test-Time Planning for World Action Models.
Before online feedback is available, \methodname{} ranks multiple future-action candidates using two
model-internal signals: \textit{flow-based video surprisal}, which estimates visual
plausibility under the WAM's own generative dynamics, and \textit{action path effort},
which measures the stability of the action denoising trajectory. After execution,
\methodname{} compares the selected imagined future with the realized observation and uses this delayed mismatch to train a lightweight online predictor. In this way, \methodname{} amortizes post-execution self-verification into pre-execution candidate
selection while keeping the backbone WAM frozen.
Our contributions are as follows:
\begin{itemize}
    \item We formulate test-time WAM control as reliability-aware candidate selection, identifying
    \emph{imagination falsifiability} as a structural advantage of causal WAMs. We show that
    sampled futures from the same prefix contain a selectable reliability gap, and that
    imagination--reality mismatch is associated with downstream failure.

    \item We introduce \methodname{}, a reward-free and verifier-free test-time planning framework for frozen
    WAMs. \methodname{} ranks best-of-$N$ future--action candidates using model-internal generative traces, namely flow-based video surprisal for visual plausibility and action path effort for action-generation stability.

    \item We propose delayed self-verification for online calibration: after execution, the realized observation audits the selected imagination, and the resulting mismatch trains a lightweight predictor that amortizes post-execution feedback into future pre-execution selection.

    \item We validate \methodname{} on RoboTwin 2.0 and a real Franka stress test. Under limited-randomization Hard RoboTwin evaluation, \methodname{} improves Hard Avg. by $+5.10$ points and Horizon-3 success by $+16.43$ points over the base WAM, while ablations show that gains come from structured selection rather than sampling more candidates alone.
\end{itemize}

\section{Related Work}
\vspace{-6pt}
\paragraph{Selection and scaling in stochastic generation.}
A broad lesson from generative modeling is that sample quality can vary substantially under the same condition, making decoding, or refinement central to deployment-time performance. 
Text-to-image systems use preference models, reward models, and inference-time refinement to select or improve generations~\cite{xu2023imagereward,kirstain2023pick,li2025reflect}. 
Other methods intervene directly in the generative trajectory, for example by amortizing test-time noise optimization or optimizing noise variables under interaction constraints~\cite{eyring2025noise,ota2025pino}. 
Our work transfers this selection view to WAM control, where the candidate is not only a visual sample but a coupled future-action rollout, and therefore must be judged by signals available before physical feedback arrives.

\vspace{-10pt}
\paragraph{Test-time guidance for robot policies.}
Generative control and offline reinforcement learning methods often guide sampled trajectories using rewards, values, constraints, or learned critics~\cite{janner2021offline,janner2022planning,wang2022diffusion}. 
Recent robot foundation-model methods similarly improve action decoding through value guidance, bidirectional sampling, external verification, or verifier-free candidate filtering~\cite{nakamoto2024steering,liu2024bidirectional,kwok2025robomonkey,jang2025verifier}. 
These approaches demonstrate the value of test-time computation, but they usually rely on additional supervision, learned evaluators, or action-level selection criteria. 
We address the complementary case where future samples are useful but uneven in reliability within WAM context, and show that selection can make test-time imagination more effective under randomized-scene uncertainty.


\section{Method}
\subsection{Preliminary: Chunked World Model and the Falsifiability Opportunity}
\label{sec:method_prelim}
We build on a causal video-action world model~\footnote{Here, ``causal'' refers to this modality-level generation order, in contrast to joint video-action formulations that denoise or model future visual and action tokens simultaneously.} that performs chunked closed-loop control.
At decision step $m$, the model conditions on the executed interaction history, represented by the causal cache $C_m=(z^{\mathrm{obs}}_{<m},a_{<m})$, where
$z^{\mathrm{obs}}_{<m}$ denotes previously realized observations encoded into the visual latent space.
Conditioned on $C_m$, the model first generates an imagined visual latent future and then decodes
the corresponding low-level action chunk:
\begin{equation}
    \hat{z}_{m,1:K} \sim q^v_\theta(\cdot \mid C_m),
    \qquad
    a_{m,1:K} \sim q^a_\theta(\cdot \mid \hat{z}_{m,1:K}, C_m),
    \qquad
    z^{\mathrm{real}}_{m,1:K}=E(o_{m,1:K}).
    \label{eq:chunked_wam}
\end{equation}
After executing the action chunk, the realized latent observation $z^{\mathrm{real}}_{m,1:K}$ is appended to the history, and the process repeats autoregressively over chunks.
This structure creates the central opportunity exploited by \methodname{}.
Before acting, the WAM commits to an explicit visual prediction $\hat{z}_{m,1:K}$; after acting, the physical world reveals $z^{\mathrm{real}}_{m,1:K}$.
Thus, the imagined future is a falsifiable prediction, auditable by the world it was meant to describe. \methodname{} uses this delayed imagination--reality discrepancy to calibrate subsequent test-time selection, while keeping the backbone WAM frozen.

Both video and action generators are implemented as flow-matching denoising processes. For clarity, the main text uses the oriented source-to-data coordinate $\tau=1-\sigma$; the native scheduler-coordinate implementation and its equivalence are given in Appendix~\ref{app:scheduler}.

\subsection{Test-Time Candidate Selection via Internal Generative Signals}
\label{sec:selection}

\begin{figure}[t]
    \centering
    \includegraphics[width=\linewidth]{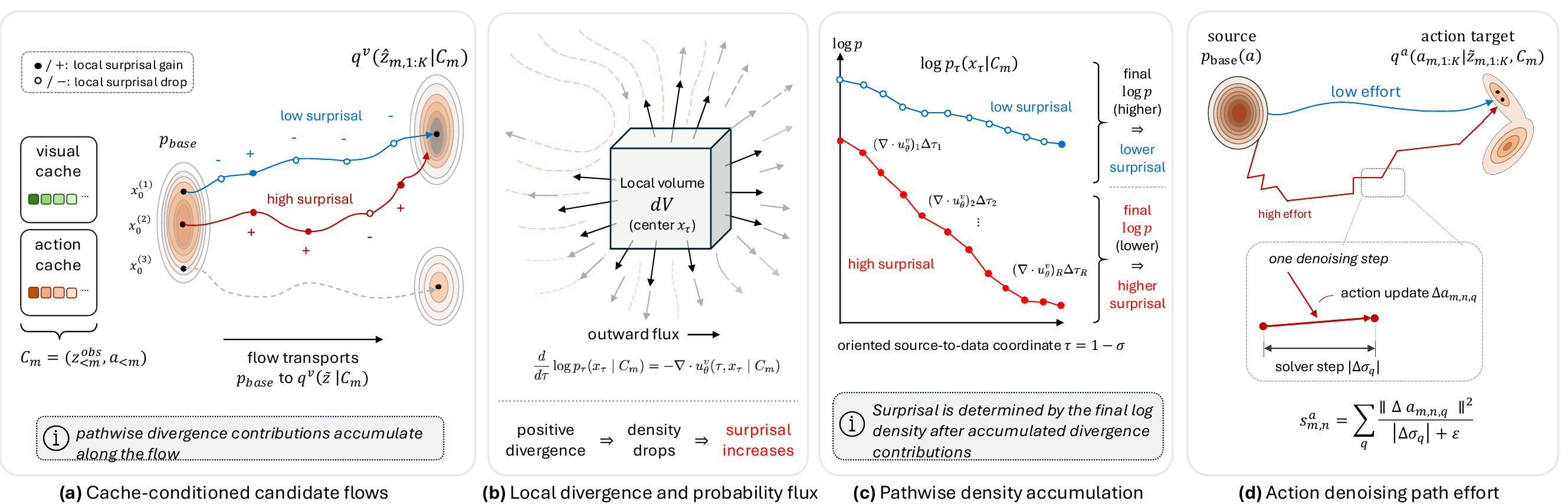}
    \caption{
    \textbf{Internal generative signals for test-time candidate selection.}
    (a) From the same cache state $C_m$, the frozen WAM samples multiple video--action candidates through stochastic conditional flows.
    (b) Positive divergence in the oriented source-to-data coordinate $\tau=1-\sigma$ corresponds to outward probability flux and local density decrease.
    (c) Accumulating divergence along the saved video solver trajectory yields the video-side surprisal score $s^{v,\mathrm{surp}}_{m,n}$.
    (d) Given an imagined future, the action branch decodes the corresponding action chunk; path effort $s^a_{m,n}$ measures normalized denoising motion in action space.
    }
    \label{fig:method_flow}
\end{figure}

The central planning problem is to choose a reliable action chunk before its real consequence is
observed. At decision step $m$, we turn the frozen WAM into a best-of-$N$ selector by sampling
\[
    \{(\hat z_{m,n,1:K}, a_{m,n,1:K})\}_{n=1}^{N},
    \qquad
    \hat z_{m,n,1:K} \sim q^v_\theta(\cdot \mid C_m),
    \quad
    a_{m,n,1:K} \sim q^a_\theta(\cdot \mid \hat z_{m,n,1:K}, C_m).
\]
All candidates share the same observation history and task conditioning; their differences therefore
reflect stochasticity in the conditional generative process rather than changes in the causal cache
(Figure~\ref{fig:method_flow}(a)). \methodname{} ranks these candidates before execution using only
generation-time traces, without rewards, external verifiers, or policy updates. We use two complementary
internal signals: video-side flow surprisal for visual plausibility and action-side path effort for action
generation stability.

\vspace{-10pt}
\paragraph{Video-side flow surprisal.}
The video branch defines a cache-conditioned future distribution over plausible visual continuations.
In a flow-matching sampler, a learned velocity field transports a base distribution toward this
conditional future distribution~\cite{lipman2022flow,liu2022flow}. As in continuous normalizing
flows~\cite{chen2018neural,grathwohl2018ffjord}, the log-density along a trajectory changes according
to the divergence of the velocity field:
\begin{equation}
    \frac{d}{d\tau}\log p_\tau(x_\tau \mid C_m)
    =
    -\nabla \cdot u^v_\theta(\tau, x_\tau \mid C_m),
    \label{eq:flow_log_density}
\end{equation}
where $\tau=1-\sigma$ is the oriented source-to-data coordinate. Intuitively, positive divergence corresponds to outward probability flux, causing local density to
drop and surprisal to increase (Figure~\ref{fig:method_flow}(b)).
Accumulating these divergence contributions along a candidate's saved trajectory determines its
final log-density: trajectories with larger positive-divergence accumulation arrive at lower-density,
higher-surprisal futures (Figure~\ref{fig:method_flow}(c)). For candidate $n$, let
$\mathcal{T}^v_{m,n}=\{(x^v_{m,n,r},\tau^v_r,\Delta\tau^v_r)\}_{r=0}^{R_v-1}$
be its saved video denoising trace. Discretizing Eq.~\eqref{eq:flow_log_density} gives the
pathwise video surprisal score:
\begin{equation}
    s^{v,\mathrm{surp}}_{m,n}
    =
    S^v_{\mathrm{base},m,n}
    +
    \sum_{r=0}^{R_v-1}
    \widehat{\nabla \cdot u^v_\theta}(\tau^v_r, x^v_{m,n,r} \mid C_m)\,
    \Delta\tau^v_r ,
    \label{eq:video_surprisal}
\end{equation}
where $S^v_{\mathrm{base},m,n}=-\log p_{\mathrm{base}}(x^v_{m,n,0})$.
We estimate the divergence with a Hutchinson trace estimator~\cite{grathwohl2018ffjord} and provide the native
$\sigma$-coordinate implementation in Appendix~\ref{app:scheduler}. Lower surprisal indicates
a higher-density imagined future under the WAM's learned visual dynamics, and is therefore preferred
before reality arrives to audit the prediction.

\vspace{-10pt}
\paragraph{Action-side path effort.}
A visually plausible future may still require an unstable low-level action trajectory. We therefore complement the video score with an action-side path-effort score derived from action denoising
updates (Figure~\ref{fig:method_flow}(d)). Let $\Delta a_{m,n,q}=x^a_{m,n,q+1}-x^a_{m,n,q}$ be the action update at denoising step $q$.
We define
\begin{equation}
    s^a_{m,n}
    =
    \sum_{q=0}^{R_a-1}
    \frac{
    \operatorname{mean}\!\left[(\Delta a_{m,n,q})^2\right]
    }{
    |\Delta\sigma^a_q|+\epsilon
    } .
    \label{eq:action_path_effort}
\end{equation}
The normalization by $|\Delta\sigma^a_q|$ measures corrective motion per unit solver step, making the score less sensitive to nonuniform scheduler spacing. Lower effort indicates a smoother and more direct action generation path, used only as a relative control-side reliability proxy among candidates sharing the same conditioning. A schedule-normalized kinetic-cost interpretation is given in Appendix~\ref{app:effort}.

\vspace{-10pt}
\paragraph{Fusion and selection.}
Since video and action scores have different scales, we normalize each score across the $N$ candidates
at the current decision step:
\begin{equation}
    \tilde{s}^{j}_{m,n}
    =
    \frac{s^{j}_{m,n}-\mu(s^{j}_{m,1:N})}
    {\operatorname{std}(s^{j}_{m,1:N})+\epsilon_s},
    \qquad j\in\{v,a\}.
    \label{eq:score_normalization}
\end{equation}
The final candidate score and selected rollout are
\begin{equation}
    c_{m,n}
    =
    \lambda_m \tilde{s}^{v}_{m,n}
    +
    (1-\lambda_m)\tilde{s}^{a}_{m,n},
    \qquad
    n_m^\star = \arg\min_n c_{m,n}.
    \label{eq:selection_rule}
\end{equation}
\methodname{} executes $a_{m,n_m^\star,1:K}$, retains the selected imagined future as visual memory, and
caches its generation features for the delayed feedback described in Section~\ref{sec:online_cal}.
Normalization edge cases and exploration are detailed in Appendix~\ref{app:scheduler}.

\begin{figure}[t]
    \centering
    \includegraphics[width=\linewidth]{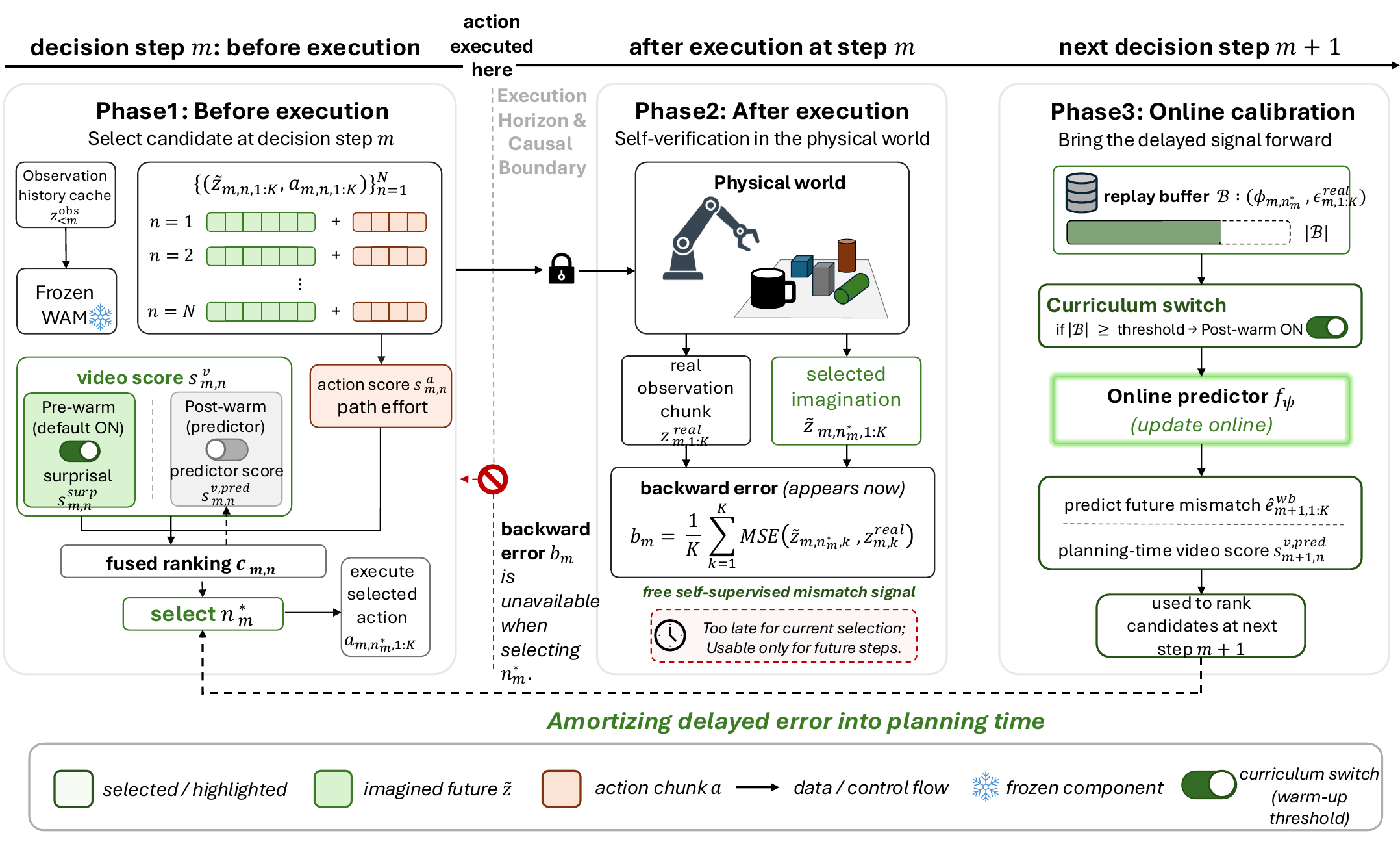}
    \caption{
    \textbf{Method overview of \methodname{}.} At each decision step $m$, the frozen WAM samples $N$ future--action candidates and selects one via pre-execution internal signals
    (\textit{Phase~1}).
    After execution, the realized observation $z^{\mathrm{real}}_{m,1:K}$ audits the selected imagination, yielding a self-supervised backward error $b_m$ that is reliable but causally delayed (\textit{Phase~2}).
    \methodname{} accumulates these mismatch labels in a replay buffer to train an online predictor $f_\psi$, amortizing the delayed signal into a planning-time video score $s^{v,\mathrm{pred}}_{m+1,n}$ for future candidate ranking (\textit{Phase~3}).
    }
    \vspace{-10pt}
    \label{fig:method_framework}
\end{figure}

\subsection{Delayed Self-Verification via Backward Error}
\label{sec:backward}
 
The most informative reliability signal arrives only after execution:
the realized observation directly audits the selected imagination.
Given the executed candidate $n_m^{\text{exec}}$, we define the
per-frame visual backward error
\begin{equation}
  e^{\text{vb}}_{m,k}
  = \operatorname{MSE}\!\left(
      \hat{z}_{m,n_m^{\text{exec}},k},\;
      z^{\text{real}}_{m,k}
    \right),
  \quad k = 1, \ldots, K,
  \label{eq:frame_bwd}
\end{equation}
and its scalar summary
$b_m = \tfrac{1}{K}\sum_k e^{\text{vb}}_{m,k}$.
The signal is entirely self-supervised—it requires no reward,
no success detector, and no external verifier.
 
We exploit $b_m$ in two ways.
First, it drives online adaptation of the video--action fusion weight.
Letting $\bar b^{\mathrm{hist}}_m$ denote the running mean of prior backward scores,
\begin{equation}
    \lambda_{m+1}
    =
    \lambda_{\mathrm{base}}
    \cdot
    \min\left(
    1,
    \frac{\bar b^{\mathrm{hist}}_m}{\max(b_m,\epsilon_b)}
    \right).
    \label{eq:lambda_update}
\end{equation}
When the current mismatch exceeds the historical baseline, trust in the video-side score is reduced
proportionally.
Second, for the executed candidate, we store a planning-time generation feature
$\phi_{m,n^{\mathrm{exec}}_m}$ together with the revealed error vector
$e^{\mathrm{vb}}_{m,1:K}$ in a replay buffer $\mathcal B$.
This links information available before execution to the ground-truth reliability revealed afterward,
forming the supervision signal for the online predictor below.

\subsection{Online Calibration: Amortizing Delayed Error into a Learned Proxy}
\label{sec:online_cal}
Flow surprisal provides a principled cold-start score, but it is computationally expensive:
for every candidate and decision step, it requires storing the denoising trace, replaying the
velocity field with gradients, and estimating divergence.
In contrast, each candidate already exposes compact generation-time features during standard video
sampling. We denote this feature vector by $\phi_{m,n}$, constructed from the denoising velocity-norm
sequence, spatially pooled future latents, and pooled frame differences. The exact construction is
given in Appendix~\ref{app:predictor_features}.

We therefore train a lightweight predictor $f_\psi$ that maps $\phi_{m,n}$ to a direct estimate of
future the backward error that will be observed after execution, namely the imagination--reality mismatch::
\begin{equation}
    \hat e^{\mathrm{vb}}_{m,n,k}
    =
    f_\psi(\phi_{m,n})_k,
    \qquad
    s^{v,\mathrm{pred}}_{m,n}
    =
    \frac{1}{K}\sum_{k=1}^{K}\hat e^{\mathrm{vb}}_{m,n,k}.
    \label{eq:predictor_score}
\end{equation}
The predictor is trained online from $\mathcal B$ via framewise regression,
\begin{equation}
    \mathcal L(\psi)
    =
    \mathbb E_{(\phi,y)\sim\mathcal B}
    \left[
    \|f_\psi(\phi)-y\|_2^2
    \right],
    \qquad
    y=e^{\mathrm{vb}}_{m,1:K}.
    \label{eq:predictor_loss}
\end{equation}
We use separate encoders for denoising dynamics and latent-motion features before a shared
prediction head; feature and architecture ablations are provided in
Appendix~\ref{app:predictor_features}.
 \vspace{-10pt}
\paragraph{Warm-up curriculum.}
\methodname{} follows a data-driven two-stage schedule illustrated in Figure~\ref{fig:method_framework}.
In the pre-warm stage, the video-side score is flow surprisal
$s^{v,\mathrm{surp}}_{m,n}$, providing a principled cold start before any deployment feedback exists.
Once $|\mathcal B|$ reaches a threshold, \methodname{} switches to post-warm: surprisal traces are discarded, divergence replay is disabled, and the video score becomes
$s^v_{m,n}=s^{v,\mathrm{pred}}_{m,n}$.
The action-side path-effort score and frozen WAM backbone remain unchanged.
Predictor weights and the replay buffer persist across episodes of the same task, while episode-local state, including the fusion weight and autoregressive cache, resets per episode.

\section{Experiments}
\subsection{Experiment Setup}
\label{sec:setup}
\paragraph{RoboTwin 2.0 simulation.}
We evaluate \methodname{} on RoboTwin 2.0~\cite{chen2025robotwin}, a 50-task bimanual manipulation benchmark covering diverse object interactions, multi-step execution, and long-horizon skills. We consider two training-randomization regimes on the same benchmark. The \emph{standard protocol} follows prior RoboTwin evaluations~\cite{li2026causal,bi2025motus}: we use the pretrained backbone trained on a mixture of 2.5k clean demonstrations and 25k heavily randomized demonstrations, and evaluate it on Easy and Hard configurations. This setting
tests whether \methodname{} can still improve a strong WAM in a near-saturated regime.

We further evaluate a \emph{limited-randomization protocol}, a harder point on the same randomization axis. Motivated by the practical setting where clean demonstrations are abundant but heavily randomized demonstrations are costly, the base WAM is first trained for 18k steps on clean scenes, then continued for 18k steps with a 70\% clean / 30\% randomized mixture. The task set, backbone architecture, and Hard evaluation remain unchanged; only the amount of randomized-scene supervision is reduced. This creates the predicted regime where the WAM has learned feasible visual-action structure, but its sampled futures remain uneven in reliability, leaving more headroom for candidate selection. Unless otherwise specified, we report average task success rate over multiple trials.~\ref{app:hard_protocol}

\begin{figure}[t]
    \centering
    \includegraphics[width=\linewidth]{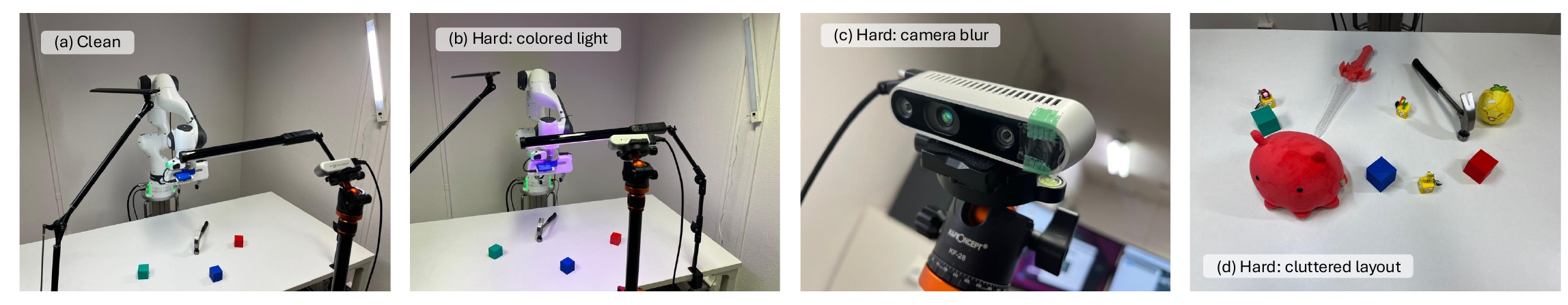}
    \caption{
    \textbf{Real-robot Franka evaluation.}
    We train on the clean hammering setup and test on unseen colored-light, camera-blur, and cluttered-layout variants.
    }
    \label{fig:franka_setup}
\end{figure}
\vspace{-10pt}
\paragraph{Real-robot Franka evaluation.}
We further evaluate \methodname{} on a real Franka tabletop hammering task, where the robot must strike the target object specified by a language instruction. 
The policy is trained only on clean demonstrations collected in a standard workspace with normal lighting and an external RGB camera. 
At test time, we keep the task interface fixed but introduce three unseen visual shifts: colored illumination, camera blur, and cluttered layouts with distractor objects (Figure~\ref{fig:franka_setup}). 
These variants directly perturb the visual evidence used by the WAM while preserving the language-conditioned control objective, creating a challenging robustness test for test-time candidate selection. 
All real-robot evaluations use the same frozen backbone; no hard variant is used for additional training or finetuning. Additional protocol details are provided in Appendix~\ref{app:franka_training_details}.

\subsection{Main Results}
\paragraph{Imagination error is behaviorally meaningful.}
\label{sec:error_sucess_failure_connection}
\begin{figure}[t]
    \centering
    \includegraphics[width=0.9\linewidth]{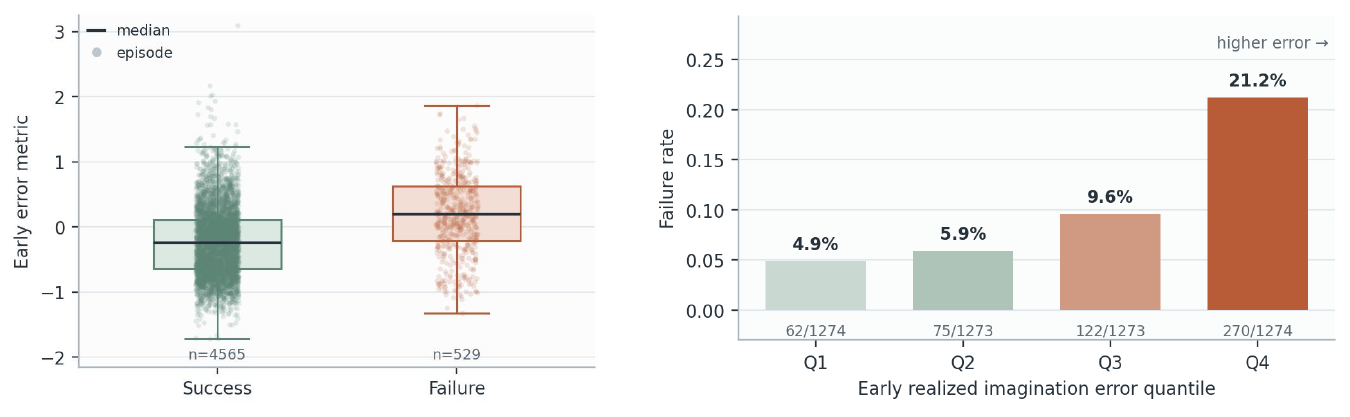}
    \caption{
    \textbf{Early imagination--reality mismatch is associated with rollout failure.}
    We aggregate task- and depth-normalized visual-back error over the first three executed chunks. Failure rate increases from $4.9\%$ in the lowest early-error quartile to $21.2\%$ in the highest.
    }
    \vspace{-13pt}
    \label{fig:error_success_failure_connection}
\end{figure}

Figure~\ref{fig:hook_pilot} shows that, under the same observation prefix, sampled imaginations can differ substantially in hindsight consistency. 
Figure~\ref{fig:error_success_failure_connection} further asks whether this mismatch is behaviorally meaningful during online rollouts, where the selected candidate is executed and its imagined latent future can be compared with the subsequently observed latent chunk.

We aggregate the mismatch over the first three decision chunks of each episode, using task- and depth-normalized log error to control for scale differences across tasks and rollout stages. 
The results show two complementary patterns: failed episodes exhibit larger early mismatch than successful ones, and failures concentrate in high-error rollouts. 
Quantitatively, the failure rate increases monotonically from $4.9\%$ to $21.2\%$ across early-error quartiles. 
Thus, the backward error used by \methodname{} is behaviorally aligned with task-level failure, making it a meaningful self-supervised target for test-time reliability estimation.
\vspace{-10pt}
\paragraph{Standard setting with a strong pretrained WAM.}
We first plug \methodname{} into the pretrained LingBot-VA backbone~\cite{li2026causal}, keeping all backbone weights frozen. As shown in Table~\ref{tab:robotwin_horizon}, \methodname{} remains competitive across horizons and improves overall Hard success from $91.55\%$ to $92.28\%$. This result shows that reliability-aware candidate selection can still provide gains on top of a strong WAM, but the absolute improvement is naturally limited by saturation: several RoboTwin tasks are already near-perfect under the fully trained standard protocol. This motivates the limited-randomization protocol below, where randomized-scene uncertainty leaves more room for test-time selection to matter.
\begin{table}[t]
\centering
\footnotesize
\setlength{\tabcolsep}{3.2pt}
\renewcommand{\arraystretch}{1.04}
\caption{
Standard RoboTwin~2.0 evaluation with the pretrained LingBot-VA backbone kept frozen. Best results are bolded and second-best results are underlined.
}
\label{tab:robotwin_horizon}
\begin{tabular}{lcccccccc}
\toprule
\multirow{2}{*}{Method}
& \multicolumn{2}{c}{H1}
& \multicolumn{2}{c}{H2}
& \multicolumn{2}{c}{H3}
& \multicolumn{2}{c}{Overall} \\
\cmidrule(lr){2-3}
\cmidrule(lr){4-5}
\cmidrule(lr){6-7}
\cmidrule(lr){8-9}
& Easy & Hard
& Easy & Hard
& Easy & Hard
& Easy & Hard \\
\midrule
X-VLA$^{*}$
& 81.6 & 82.5
& 59.3 & 55.9
& 61.2 & 66.0
& 72.9 & 72.8 \\

$\pi_0$
& 66.5 & 61.6
& 66.1 & 54.7
& 61.6 & 50.2
& 65.9 & 58.4 \\

$\pi_{0.5}$
& 85.1 & 80.2
& 79.3 & 73.0
& 78.6 & 67.4
& 82.7 & 76.8 \\

Motus
& 91.0 & 90.6
& 85.2 & 80.9
& 85.0 & 84.2
& 88.7 & 87.0 \\

LingBot-VA
& \underline{94.2} & \underline{93.6}
& \textbf{90.3} & \underline{86.9}
& \underline{93.2} & \textbf{93.2}
& \underline{92.9} & \underline{91.6} \\

\methodname{}
& \textbf{94.5} & \textbf{93.6}
& \underline{89.9} & \textbf{89.4}
& \textbf{93.3} & \underline{93.2}
& \textbf{93.0} & \textbf{92.3} \\
\bottomrule
\end{tabular}
\end{table}
\vspace{-10pt}
\paragraph{Generalization under limited randomized-scene supervision.}
\begin{wraptable}{r}{0.50\linewidth}
\vspace{-8pt}
\centering
\footnotesize
\setlength{\tabcolsep}{4.5pt}
\renewcommand{\arraystretch}{1.04}
\caption{
Limited-randomization Hard results. $\Delta$ denotes absolute gain.
}
\label{tab:robotwin_hard}
\begin{tabular}{lcccc}
\toprule
Method & H1 & H2 & H3 & Avg. \\
\midrule
LingBot-VA
& 63.33 & 49.33 & 30.00 & 55.80 \\
\methodname{}
& \textbf{66.72} & \textbf{53.69} & \textbf{46.43} & \textbf{60.90} \\
$\Delta$
& \textcolor{darkgreen}{+3.39}
& \textcolor{darkgreen}{+4.36}
& \textcolor{darkgreen}{+16.43}
& \textcolor{darkgreen}{+5.10} \\
\bottomrule
\end{tabular}
\vspace{-8pt}
\end{wraptable}
Table~\ref{tab:robotwin_hard} evaluates the same selection mechanism under the less saturated limited-randomization protocol. Here, the base WAM has seen randomized scenes during training, but not enough to fully master their visual and geometric variation. This directly targets our intended failure mode: sampled futures can be plausible under the model, yet still vary substantially in reliability under randomized test scenes.

\methodname{} improves Hard success from $55.80\%$ to $60.90\%$ without updating the backbone. The gain is especially large on Horizon-3 tasks, where success increases by $+16.43$ points. This supports the core hypothesis that test-time candidate selection is most useful when the base model has learned enough structure to imagine feasible futures, but its stochastic samples remain unreliable enough that selecting a better future-action rollout can prevent early errors from compounding.
\vspace{-10pt}
\paragraph{Real-robot Franka stress test.}
Figure~\ref{fig:franka_setup} and the supplementary rollout videos provide a qualitative real-robot stress test under visual shifts not seen during clean-data training. 
In the colored-light case, both the frozen base WAM and the $\pi0.5$ baseline fail to complete the hammering task, while \methodname{} selects an executable trajectory. 
These results provide preliminary evidence that test-time selection can recover more reliable actions from the same frozen WAM under shifted or degraded visual evidence.

\subsection{Ablation Study}
\label{sec:ablation}

We ablate \methodname{} under the hard generalization protocol in Section~\ref{sec:setup}, focusing on module contributions, candidate count~$N$, and predictor design.
\vspace{-10pt}
\paragraph{Module ablation.}
\begin{table}[t]
\centering
\small
\caption{%
    Module ablation on RoboTwin~2.0.
    ``Video'' and ``Action'' denote the video-side and action-side
    scoring branches; ``Online Pred.''\ denotes the online reliability
    predictor. \emph{Random-$N$} selects uniformly among $N$ candidates
    and serves as the baseline.
}
\label{tab:ablation_module}
\setlength{\tabcolsep}{0pt}
\renewcommand{\arraystretch}{1.05}
\begin{tabular*}{\linewidth}{@{}l@{\extracolsep{\fill}}cccccccc@{}}
\toprule
\multirow{2}{*}{Variant}
  & \multirow{2}{*}{\makecell[c]{Video\\Branch}}
  & \multirow{2}{*}{\makecell[c]{Action\\Branch}}
  & \multirow{2}{*}{\makecell[c]{Online\\Pred.}}
  & \multirow{2}{*}{\makecell[c]{Clean\\Avg.~(\%)}}
  & \multicolumn{3}{c}{Hard horizon-wise (\%)}
  & \multirow{2}{*}{\makecell[c]{Hard\\Avg.~(\%)}} \\
\cmidrule(lr){6-8}
& & & & & H\,=\,1 & H\,=\,2 & H\,=\,3 & \\
\midrule
Random-$N$
  & \xmark & \xmark & \xmark
  & 91.1 & 62.2 & 51.1 & 27.0 & 53.5 \\
Action-only
  & \xmark & Path Eff. & \xmark
  & \textbf{93.3} & 64.9 & 55.2 & 21.6 & 55.1 \\
Surprisal-only
  & Surp. & \xmark & \xmark
  & 90.4 & 64.9 & \textbf{56.5} & 23.0 & 55.4 \\
Predictor-only
  & Pred. & \xmark & \cmark
  & 92.9 & \textbf{68.9} & 55.7 & 35.4 & 59.3 \\
\midrule
\textbf{WCD} (ours)
  & Surp.$\!\to\!$Pred. & Path Eff. & \cmark
  & 92.3 & 66.7 & 53.7 & \textbf{46.4} & \textbf{60.9} \\
\bottomrule
\end{tabular*}
\end{table}
Table~\ref{tab:ablation_module} shows that sampling alone is insufficient: Random-$N$ reaches $91.1\%$ on clean scenes but only $53.5\%$ on Hard. The single-branch variants reveal complementary failure modes. Path effort improves short-horizon robustness, surprisal filters implausible visual futures under distribution shift, and the predictor gives the best Horizon-3 result among the ablations, but none is reliable alone. Combining them yields the best Hard average ($60.90\%$), improving over Random-$N$ by $+7.40$ and over the strongest single-branch variant by $+5.50$. This supports the need to combine video plausibility, action stability, and delayed self-verification.
\vspace{-10pt}
\paragraph{Effect of candidate count~$N$.}
\begin{figure}[t]
    \centering
    \includegraphics[width=\linewidth]{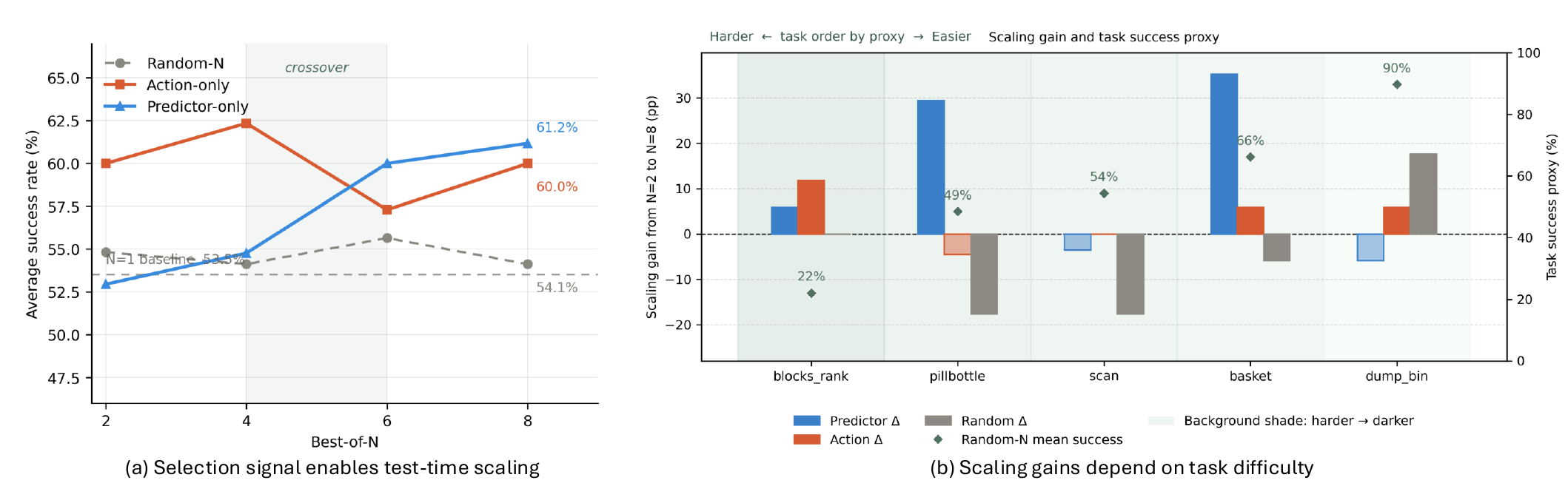}
    \caption{
    \textbf{Test-time scaling requires structured selection.}
    (a) More candidates improve success only when paired with reliable selection.
    (b) Scaling gains concentrate on tasks with remaining headroom.
    }
    \vspace{-10pt}
    \label{fig:bon_scaling}
\end{figure}

Figure~\ref{fig:bon_scaling} shows that test-time scaling only helps when additional candidates can be ranked. At $N{=}1$, all methods reduce to the same single rollout. As $N$ increases, Random-$N$ fluctuates without a consistent trend, whereas structured selectors improve more reliably: action-only peaks at smaller pools, while predictor-only benefits more at larger pools ($N{=}6$--$8$). The task-level view in Figure~\ref{fig:bon_scaling}(b) further shows that gains are largest on tasks with remaining headroom and shrink on saturated tasks such as \texttt{dump\_bin\_bigbin}. More candidates are therefore useful only when paired with a reliability signal that can expose better futures.

\section{Conclusion}

We presented \methodname{}, a reward-free and verifier-free test-time planning framework that treats WAM
imagination as a set of falsifiable future--action hypotheses rather than a one-shot intermediate
generation. By combining flow-based video surprisal, action path effort, and an online predictor
trained from delayed imagination--reality mismatch, \methodname{} selects and calibrates reliable candidates
while keeping the backbone WAM frozen.

Experiments show that this reliability-aware selection is most useful when stochastic futures remain
uncertain. Early imagination--reality mismatch is associated with rollout failure, structured selectors
outperform random sampling, and under limited-randomization Hard RoboTwin evaluation \methodname{}
improves success from $55.80\%$ to $60.90\%$, with a $+16.43$ point gain on Horizon-3 tasks.
Real Franka stress tests further suggest that the same selection mechanism can recover executable
actions under visual shifts.

A remaining trade-off is test-time compute: \methodname{} improves robustness by spending additional parallel candidate generation.
Appendix~\ref{app:latency} shows that the costly flow-surprisal replay is only at cold-start stage, while the post-warm online predictor adds negligible scoring overhead compared with WAM generation.
Future work should study faster backbone WAM generation and optimized deployment pipelines for high-frequency robot control.



\newpage
\bibliographystyle{plainnat}
\bibliography{references}

@article{li2026causal,
  title={Causal World Modeling for Robot Control},
  author={Li, Lin and Zhang, Qihang and Luo, Yiming and Yang, Shuai and Wang, Ruilin and Han, Fei and Yu, Mingrui and Gao, Zelin and Xue, Nan and Zhu, Xing and others},
  journal={arXiv preprint arXiv:2601.21998},
  year={2026}
}

@misc{ye2026worldactionmodelszeroshot,
      title={World Action Models are Zero-shot Policies}, 
      author={Seonghyeon Ye and Yunhao Ge and Kaiyuan Zheng and Shenyuan Gao and Sihyun Yu and George Kurian and Suneel Indupuru and You Liang Tan and Chuning Zhu and Jiannan Xiang and Ayaan Malik and Kyungmin Lee and William Liang and Nadun Ranawaka and Jiasheng Gu and Yinzhen Xu and Guanzhi Wang and Fengyuan Hu and Avnish Narayan and Johan Bjorck and Jing Wang and Gwanghyun Kim and Dantong Niu and Ruijie Zheng and Yuqi Xie and Jimmy Wu and Qi Wang and Ryan Julian and Danfei Xu and Yilun Du and Yevgen Chebotar and Scott Reed and Jan Kautz and Yuke Zhu and Linxi "Jim" Fan and Joel Jang},
      year={2026},
      eprint={2602.15922},
      archivePrefix={arXiv},
      primaryClass={cs.RO},
      url={https://arxiv.org/abs/2602.15922}, 
}

@article{kim2026cosmos,
  title={Cosmos policy: Fine-tuning video models for visuomotor control and planning},
  author={Kim, Moo Jin and Gao, Yihuai and Lin, Tsung-Yi and Lin, Yen-Chen and Ge, Yunhao and Lam, Grace and Liang, Percy and Song, Shuran and Liu, Ming-Yu and Finn, Chelsea and others},
  journal={arXiv preprint arXiv:2601.16163},
  year={2026}
}

@article{ye2026gigaworld,
  title={GigaWorld-Policy: An Efficient Action-Centered World--Action Model},
  author={Ye, Angen and Wang, Boyuan and Ni, Chaojun and Huang, Guan and Zhao, Guosheng and Li, Hao and Li, Hengtao and Li, Jie and Lv, Jindi and Liu, Jingyu and others},
  journal={arXiv preprint arXiv:2603.17240},
  year={2026}
}

@inproceedings{hu2025video,
  title={Video Prediction Policy: A Generalist Robot Policy with Predictive Visual Representations},
  author={Hu, Yucheng and Guo, Yanjiang and Wang, Pengchao and Chen, Xiaoyu and Wang, Yen-Jen and Zhang, Jianke and Sreenath, Koushil and Lu, Chaochao and Chen, Jianyu},
  booktitle={International Conference on Machine Learning},
  pages={24328--24346},
  year={2025},
  organization={PMLR}
}

@article{janner2021offline,
  title={Offline reinforcement learning as one big sequence modeling problem},
  author={Janner, Michael and Li, Qiyang and Levine, Sergey},
  journal={Advances in neural information processing systems},
  volume={34},
  pages={1273--1286},
  year={2021}
}

@article{janner2022planning,
  title={Planning with diffusion for flexible behavior synthesis},
  author={Janner, Michael and Du, Yilun and Tenenbaum, Joshua B and Levine, Sergey},
  journal={arXiv preprint arXiv:2205.09991},
  year={2022}
}

@article{nakamoto2024steering,
  title={Steering your generalists: Improving robotic foundation models via value guidance},
  author={Nakamoto, Mitsuhiko and Mees, Oier and Kumar, Aviral and Levine, Sergey},
  journal={arXiv preprint arXiv:2410.13816},
  year={2024}
}

@article{kwok2025robomonkey,
  title={Robomonkey: Scaling test-time sampling and verification for vision-language-action models},
  author={Kwok, Jacky and Agia, Christopher and Sinha, Rohan and Foutter, Matt and Li, Shulu and Stoica, Ion and Mirhoseini, Azalia and Pavone, Marco},
  journal={arXiv preprint arXiv:2506.17811},
  year={2025}
}

@article{liu2024bidirectional,
  title={Bidirectional decoding: Improving action chunking via guided test-time sampling},
  author={Liu, Yuejiang and Hamid, Jubayer Ibn and Xie, Annie and Lee, Yoonho and Du, Maximilian and Finn, Chelsea},
  journal={arXiv preprint arXiv:2408.17355},
  year={2024}
}

@article{jang2025verifier,
  title={Verifier-free Test-Time Sampling for Vision Language Action Models},
  author={Jang, Suhyeok and Kim, Dongyoung and Kim, Changyeon and Kim, Youngsuk and Shin, Jinwoo},
  journal={arXiv preprint arXiv:2510.05681},
  year={2025}
}

@article{chen2025robotwin,
  title={Robotwin 2.0: A scalable data generator and benchmark with strong domain randomization for robust bimanual robotic manipulation},
  author={Chen, Tianxing and Chen, Zanxin and Chen, Baijun and Cai, Zijian and Liu, Yibin and Li, Zixuan and Liang, Qiwei and Lin, Xianliang and Ge, Yiheng and Gu, Zhenyu and others},
  journal={arXiv preprint arXiv:2506.18088},
  year={2025}
}

@article{kirstain2023pick,
  title={Pick-a-pic: An open dataset of user preferences for text-to-image generation},
  author={Kirstain, Yuval and Polyak, Adam and Singer, Uriel and Matiana, Shahbuland and Penna, Joe and Levy, Omer},
  journal={Advances in neural information processing systems},
  volume={36},
  pages={36652--36663},
  year={2023}
}

@inproceedings{li2025reflect,
  title={Reflect-dit: Inference-time scaling for text-to-image diffusion transformers via in-context reflection},
  author={Li, Shufan and Kallidromitis, Konstantinos and Gokul, Akash and Koneru, Arsh and Kato, Yusuke and Kozuka, Kazuki and Grover, Aditya},
  booktitle={Proceedings of the IEEE/CVF International Conference on Computer Vision},
  pages={15657--15668},
  year={2025}
}

@article{xu2023imagereward,
  title={Imagereward: Learning and evaluating human preferences for text-to-image generation},
  author={Xu, Jiazheng and Liu, Xiao and Wu, Yuchen and Tong, Yuxuan and Li, Qinkai and Ding, Ming and Tang, Jie and Dong, Yuxiao},
  journal={Advances in Neural Information Processing Systems},
  volume={36},
  pages={15903--15935},
  year={2023}
}

@article{kim2024openvla,
  title={Openvla: An open-source vision-language-action model},
  author={Kim, Moo Jin and Pertsch, Karl and Karamcheti, Siddharth and Xiao, Ted and Balakrishna, Ashwin and Nair, Suraj and Rafailov, Rafael and Foster, Ethan and Lam, Grace and Sanketi, Pannag and others},
  journal={arXiv preprint arXiv:2406.09246},
  year={2024}
}

@article{black2024pi0,
  title={{$\pi_0$: A Vision-Language-Action Flow Model for General Robot Control}},
  author={Black, Kevin and Brown, Noah and Driess, Danny and Esmail, Adnan and Equi, Michael and Finn, Chelsea and Fusai, Niccolo and Groom, Lachy and Hausman, Karol and Ichter, Brian and others},
  journal={arXiv preprint arXiv:2410.24164},
  year={2024}
}

@inproceedings{black2025pi05,
  title     = {$\pi_{0.5}$: A Vision-Language-Action Model with Open-World Generalization},
  author    = {Black, Kevin and Brown, Noah and Darpinian, James and Dhabalia, Karan
               and Driess, Danny and Esmail, Adnan and Equi, Michael Robert
               and Finn, Chelsea and Fusai, Niccolo and Galliker, Manuel Y.
               and Ghosh, Dibya and Groom, Lachy and Hausman, Karol
               and ichter, brian and Jakubczak, Szymon and Jones, Tim
               and Ke, Liyiming and LeBlanc, Devin and Levine, Sergey
               and Li-Bell, Adrian and Mothukuri, Mohith and Nair, Suraj
               and Pertsch, Karl and Ren, Allen Z. and Shi, Lucy Xiaoyang
               and Smith, Laura and Springenberg, Jost Tobias and Stachowicz, Kyle
               and Tanner, James and Vuong, Quan and Walke, Homer and Walling, Anna
               and Wang, Haohuan and Yu, Lili and Zhilinsky, Ury},
  booktitle = {Proceedings of The 9th Conference on Robot Learning},
  series    = {Proceedings of Machine Learning Research},
  volume    = {305},
  pages     = {17--40},
  year      = {2025},
  publisher = {PMLR}
}

@article{eyring2025noise,
  title={Noise hypernetworks: Amortizing test-time compute in diffusion models},
  author={Eyring, Luca and Karthik, Shyamgopal and Dosovitskiy, Alexey and Ruiz, Nataniel and Akata, Zeynep},
  journal={arXiv preprint arXiv:2508.09968},
  year={2025}
}

@inproceedings{ota2025pino,
  title={Pino: Person-interaction noise optimization for long-duration and customizable motion generation of arbitrary-sized groups},
  author={Ota, Sakuya and Yu, Qing and Fujiwara, Kent and Ikehata, Satoshi and Sato, Ikuro},
  booktitle={Proceedings of the IEEE/CVF International Conference on Computer Vision},
  pages={10676--10685},
  year={2025}
}

@article{chen2018neural,
  title={Neural ordinary differential equations},
  author={Chen, Ricky TQ and Rubanova, Yulia and Bettencourt, Jesse and Duvenaud, David K},
  journal={Advances in neural information processing systems},
  volume={31},
  year={2018}
}

@article{grathwohl2018ffjord,
  title={Ffjord: Free-form continuous dynamics for scalable reversible generative models},
  author={Grathwohl, Will and Chen, Ricky TQ and Bettencourt, Jesse and Sutskever, Ilya and Duvenaud, David},
  journal={arXiv preprint arXiv:1810.01367},
  year={2018}
}

@article{lipman2022flow,
  title={Flow matching for generative modeling},
  author={Lipman, Yaron and Chen, Ricky TQ and Ben-Hamu, Heli and Nickel, Maximilian and Le, Matt},
  journal={arXiv preprint arXiv:2210.02747},
  year={2022}
}

@article{bi2025motus,
  title={Motus: A unified latent action world model},
  author={Bi, Hongzhe and Tan, Hengkai and Xie, Shenghao and Wang, Zeyuan and Huang, Shuhe and Liu, Haitian and Zhao, Ruowen and Feng, Yao and Xiang, Chendong and Rong, Yinze and others},
  journal={arXiv preprint arXiv:2512.13030},
  year={2025}
}

@article{tian2024predictive,
  title={Predictive inverse dynamics models are scalable learners for robotic manipulation},
  author={Tian, Yang and Yang, Sizhe and Zeng, Jia and Wang, Ping and Lin, Dahua and Dong, Hao and Pang, Jiangmiao},
  journal={arXiv preprint arXiv:2412.15109},
  year={2024}
}

@article{liu2022flow,
  title={Flow straight and fast: Learning to generate and transfer data with rectified flow},
  author={Liu, Xingchao and Gong, Chengyue and Liu, Qiang},
  journal={arXiv preprint arXiv:2209.03003},
  year={2022}
}

@article{wang2022diffusion,
  title={Diffusion policies as an expressive policy class for offline reinforcement learning},
  author={Wang, Zhendong and Hunt, Jonathan J and Zhou, Mingyuan},
  journal={arXiv preprint arXiv:2208.06193},
  year={2022}
}

@article{du2023learning,
  title={Learning universal policies via text-guided video generation},
  author={Du, Yilun and Yang, Sherry and Dai, Bo and Dai, Hanjun and Nachum, Ofir and Tenenbaum, Josh and Schuurmans, Dale and Abbeel, Pieter},
  journal={Advances in neural information processing systems},
  volume={36},
  pages={9156--9172},
  year={2023}
}

@article{ebert2018visual,
  title={Visual foresight: Model-based deep reinforcement learning for vision-based robotic control},
  author={Ebert, Frederik and Finn, Chelsea and Dasari, Sudeep and Xie, Annie and Lee, Alex and Levine, Sergey},
  journal={arXiv preprint arXiv:1812.00568},
  year={2018}
}

@inproceedings{srinivas2018universal,
  title={Universal planning networks: Learning generalizable representations for visuomotor control},
  author={Srinivas, Aravind and Jabri, Allan and Abbeel, Pieter and Levine, Sergey and Finn, Chelsea},
  booktitle={International conference on machine learning},
  pages={4732--4741},
  year={2018},
  organization={PMLR}
}

@article{fragkiadaki2015learning,
  title={Learning visual predictive models of physics for playing billiards},
  author={Fragkiadaki, Katerina and Agrawal, Pulkit and Levine, Sergey and Malik, Jitendra},
  journal={arXiv preprint arXiv:1511.07404},
  year={2015}
}


\newpage

\end{document}